%% file: main.tex
\documentclass[runningheads]{llncs}

\usepackage{eccv}

\usepackage{eccvabbrv}

\usepackage{graphicx}
\usepackage{booktabs}
\usepackage{multirow}
\usepackage{multicol}
\usepackage{comment}
\usepackage{tikz,pgfplots}

\usepackage[accsupp]{axessibility}  

\usepackage[pagebackref,breaklinks,colorlinks]{hyperref}

\usepackage{orcidlink}

\begin{document}


\title{Interpretable Action Recognition on Hard to Classify Actions}


\author{Anastasia Anichenko\inst{1}\and
Frank Guerin\inst{1}\and
Andrew Gilbert\inst{1}}

\authorrunning{A.~Anichenko et al.}

\institute{University of Surrey, Guildford, UK
\\
\email{\{f.guerin,a.gilbert\}@surrey.ac.uk}}

\maketitle

\begin{abstract}
We investigate a human-like interpretable model of video understanding.
Humans recognise  complex activities in video  by recognising critical spatio-temporal relations among explicitly recognised objects and parts, for example, an object entering the aperture of a container. 
To mimic this we build on a model which uses positions of objects and hands, and their motions, to recognise the activity taking place. 
To improve this model we focussed on three of the most confused classes (for this model) and identified that the lack of 3D information was the major problem. 
To address this we extended our basic model by adding  3D awareness in two ways:~(1) A state-of-the-art object detection
model was fine-tuned to determine the difference between “Container” and “NotContainer” in
order to  integrate object shape information into the existing object features. (2) A state-of-the-art depth estimation model was used to extract depth values for individual objects and
calculate depth relations to expand the existing relations used our interpretable model. These 3D extensions to our basic model were evaluated on a subset of three superficially similar “Putting”
actions from the Something-Something-v2 dataset. The results showed that the container detector did not improve performance, but the addition of depth relations made a significant improvement to performance.


\end{abstract}


\section{Introduction}
\label{sec:intro}
\input{tex/intro.tex}

\section{The Original ``Top Down Model'' (TDM)}
\label{sec:method}
\input{tex/method}

\section{Results}
\label{sec:exp} 
\input{tex/results}

\section{Discussion and Conclusion}
\label{sec:conclusion} 
\input{tex/conclusion}

\bibliographystyle{splncs04}
\bibliography{egbib,refs}
\end{document}

%% file: tex/intro.tex
There are a number of motivations for investigating a human-like interpretable approach to video activity recognition. Firstly we may aim to improve the performance of automated activity recognition, since humans are still able to beat computers at this task, and a human-like approach may bring us closer to human performance. Secondly, interpretable models may be desirable in AI systems where human society has an expectation that a decision can be explained, and justified. For example if a member of a minority ethnic group is flagged for suspicious activity in a surveillance video, and questioned by police as a result, there is an expectation that the decision process of the AI system can be inspected to ensure that it is not applying racial bias. Interpretability also has advantages when systems need to be debugged, because error analysis can give human-understandable reasons for the failure, and these can be acted upon to fix the problem. Thirdly, if we want to understand more about the human visual system, and how it interprets video, it would be insightful to be able to build models of its behaviour, and through iterative experimental comparisons, to bring those models closer and closer to matching the behaviour of the human visual system.


Research has shown \cite{BenYosef2018ImageIA} that humans discriminate similar activities by differences in relationships among critical parts of objects, such as the position of the hand relative to another person's back, differentiating between fighting and hugging. While such relationships could, in principle, be learnt by a deep learning approach, in practice, deep learning tends not to make use of features that are easily interpretable by humans, as the training process would need to be conditioned to take human characteristics into account. 


%% file: tex/method.tex
\begin{figure}[htbp]
\centering
\includegraphics[width=\textwidth]{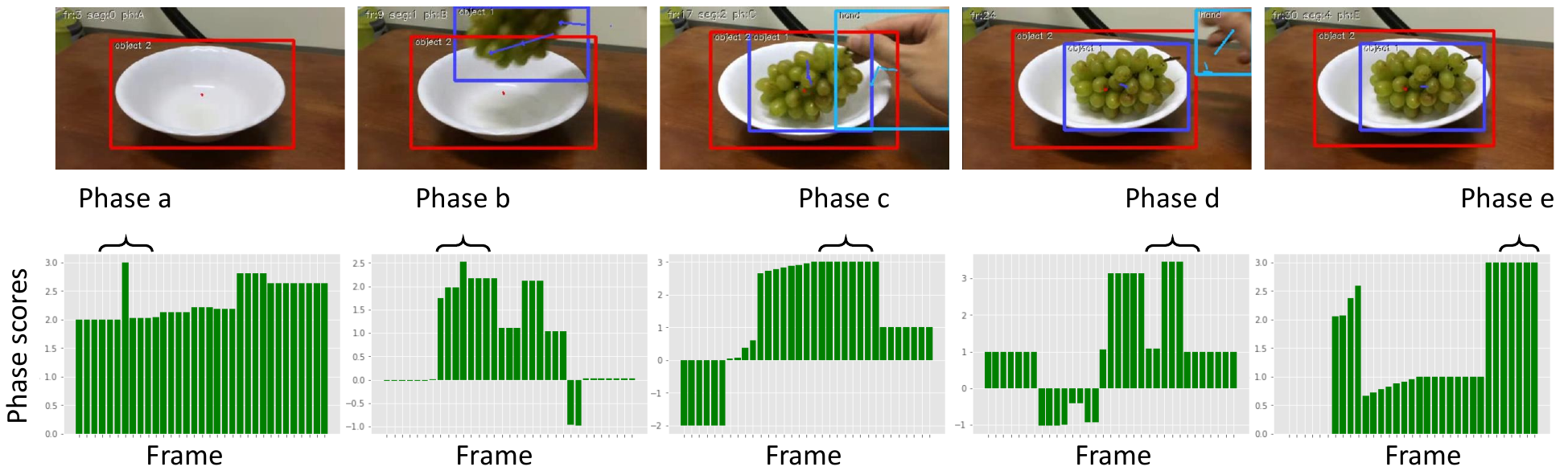}
\caption{Activity frames illustrating the 5 phases for the action 'Putting something into something'. (red and blue bounding boxes indicate objects and light blue the hand) \\
}
\label{fig:phase_score}
\end{figure}

The high-level overview of the  system we build on  is as follows (note that this description is brief as more detail is in the source \cite{TDMref}):
We use as an input only the bounding boxes of the principal objects and hands from the video frames (there are no lower level visual or optical flow features from the videos). 
We interpret the video as a possible member of each of the possible action categories using an action specific model for each class. I.e. 
every single model is attempted on the video, to see which produces a best fit.
This is seen as a top-down approach because each model has a prior conception of what a certain activity should be, and tries to impose this on the data to see if there is a fit.
The first step in applying a model is to temporally segment the video into five `phases' for each action category, representing the sequential stages involved in that action. 
The phases are labelled $a,b,c,d$ or $e$ and a typical segmentation is illustrated in Fig.~\ref{fig:phase_score} (top).

\begin{itemize}
\item Phase $a$: The object(s) is present in the scene, the manipulation has not happened;

\item Phase $b$: The hand enters (possibly carrying an object);

\item Phase $c$: The critical manipulation happens (e.g. object placed or picked);

\item Phase $d$: The hand departs (possibly carrying an object);

\item Phase $e$: The objects are present, with the result of the manipulation evident.
\end{itemize}

Most videos depict all five phases, but some do not; e.g. some videos start at phase $c$, with the hand already contacting the moved object, while other videos finish at $c$ without sufficient frames to make a clear $d$. 

Each action specific model has a method of assigning phases, which is learned from a small (approx. 25) number of labelled examples.
Once phases are assigned, we then compute feature vectors characterising each phase.
Feature vectors include data about relations among bounding boxes of the two principal objects and the hands, for example, object size, object movement since previous frame, relative movement between two objects, object moving with the hand, or moving relative to the hand, etc. 
Using the computed feature vectors we  train a random forest classifier for each category (with the positive examples being that category and other categories being negative examples). When doing multi-class classification, the highest probability random forest prediction is returned as the class.

\section{Adding 3D Awareness}
Firstly a container detection model was implemented to augment the features for the objects. Detectron2 was used as the base, and was the fine-tuned using the OpenImagesv6 dataset, which was reclassified as container or not container.
The best performance that could be achieved reached an
accuracy of 69.0\% on a validation set. Based on the error analysis performed, the model appears
to be classifying the underlying classes that were merged into "Container" and "NotContainer"
and failing to generalize on common features that all the samples shared. This is an indication
that within the scope of experimentation performed Detectron2 simply does not have the capacity
to learn 3D features such as "concavity" based off of 2D properties alone.
This illustrates the general phenomenon that it is extremely challenging to achieve a human level of performance on tasks that such as classifying affordances of everyday household objects (which have a huge variety of surface forms).

The second 3D element was depth understanding. Monocular depth estimation was performed using SharpNet \cite{DBLP:journals/corr/abs-1905-08598}. The accuracy of the depth estimation varies widely for different videos. The biggest challenge is the quality of videos and their "in the wild"
nature. Failures are evident when there is significant  motion within the
frame as well as  challenging angles of view which may differ from the perspectives the model was trained on. The estimated depth was added as the four following features: Depth of object 1,
Depth of object 2, Depth of hand, Depth difference between onject 1 and object 2.

%% file: tex/results.tex
To assess the performance of our approach, we used  Something-Something V2 \cite{goyal2017something} (SSV2) to evaluate the action recognition  task.
All models are evaluated on the following subset of actions from SSV2:
\begin{itemize}
    \item  106 : "Putting something into something"
    \item 112 : "Putting something onto something"
    \item 118 : "Putting something underneath something"
\end{itemize}

When evaluating all models the original TDM (Top-Down Model \cite{TDMref}) is used as a baseline. The results of all models trained can be seen in Table \ref{tab:results}. 

\begin{table}[ht]
    \centering
    \begin{tabular}{lcccc|cccc}
        \toprule
        Metric & \multicolumn{4}{c}{Precision} & \multicolumn{4}{c}{Recall} \\
        SSV2 class & 106 & 112 & 118 & avg & 106 & 112 & 118 & avg \\
        \midrule
        TDM (Ours) baseline \cite{TDMref} & 0.69 & \textcolor{blue}{0.47} & 0.29 & 0.48 & 0.63 & 0.59 & 0.24 & 0.49 \\
        3D CNN \cite{goyal2017something} & 0.61 & 0.36 & 0.00 & 0.32 & \textcolor{blue}{0.84} & 0.22 & 0.00 & 0.36 \\
        VideoMAE \cite{VideoMAEref} & \textbf{0.89} & \textbf{0.71} & \textbf{0.71} & \textbf{0.77} & \textbf{0.86} & \textbf{0.80} & \textbf{0.60} & \textbf{0.76} \\
        Ours + initial improvements & \textcolor{blue}{0.72} & 0.45 & 0.32 & 0.49 & 0.62 & \textcolor{blue}{0.68} & 0.14 & 0.48 \\
        Ours + container detection & 0.68 & 0.41 & 0.32 & 0.47 & 0.57 & 0.54 & \textcolor{blue}{0.34} & 0.48 \\
        Ours + depth relations & \textcolor{blue}{0.72} & 0.45 & \textcolor{blue}{0.37} & \textcolor{blue}{0.51} & 0.66 & 0.59 & 0.26 & \textcolor{blue}{0.50} \\
        Ours + container detection +& 0.66 & 0.46 & 0.30 & 0.47 & 0.71 & 0.42 & 0.24 & 0.46 \\
         depth relations \\
        \bottomrule
    \end{tabular}
    \caption{Precision and recall rates for each model tested on the validation subset. Macro average is used instead of weighted average. The highest result is highlighted in black whilst the second highest is highlighted in \textcolor{blue}{blue}.}
    \label{tab:results}
\end{table}

%% file: tex/conclusion.tex
In this work, we added 3D information both to object features (for container/ notContainer) and also to object positions. Only the latter improved performance for the system. Even then the performance we can achieve with our human-like approach falls far short of that achievable by mainstream deep learning approaches.
Some reasons for this are clear:
Our approach has very little information about the objects in the scene, for example the hand is simply described by a 2D bounding box. In contrast, a human observing a hand moving and acting in a video scene can form a pretty good estimation of its 6D pose in space (including all joint positions), and can use this to make good guesses about the grasp it is performing, and the manipulation done with an object. Similarly for the objects and context (surrounding objects): humans can describe a lot of detail. Our model is extremely impoverished compared to the level of detail a human brain extracts.
We are not aware of any work which can model the features a human extracts at a reasonable level of detail, and there are very few people in the world that seem to be even attempting this human-like approach. The most detailed approach to modelling human-like features that we are aware of is Ben Yosef et al.'s \cite{BenYosef2018ImageIA} work on still images; it shows that a lot of work needs to be done to capture human-like features even for a very small number of objects. 

In general, based on Ullman's models, human vision seems to use fewer layers of processing than deep learning, but the relations and features extracted by a level tend to be much more complicated (requiring more advanced algorithms) than the function that a single layer of deep learning can implement. Humans use a small number of advanced and critical features and relationships to discriminate categories. Deep learning uses a very large number of `weak' relations and features. For this reason deep learning degrades gradually as the input is reduced e.g. by blurring or cropping. Human vision is quite robust to degraded input, until a threshold is reached (where a critical feature is no longer recognisable), and then it drops dramatically.

%
%